\documentclass[conference]{IEEEtran}
\IEEEoverridecommandlockouts

\usepackage{cite}
\usepackage{amsmath,amssymb,amsfonts}
\usepackage{algorithmic}
\usepackage{graphicx}
\usepackage{textcomp}
\usepackage{xcolor}
\usepackage{amssymb}
\usepackage{graphicx}
\usepackage{booktabs}
\usepackage{subcaption}
\usepackage{marginnote}
\usepackage{url}
\def\BibTeX{{\rm B\kern-.05em{\sc i\kern-.025em b}\kern-.08em
    T\kern-.1667em\lower.7ex\hbox{E}\kern-.125emX}}

\begin{document}

\title{The Role of Touch: Towards Optimal Tactile Sensing Distribution in Anthropomorphic Hands\\for Dexterous In-Hand Manipulation\\
\thanks{This work was supported by the LARSyS FCT funding (DOI: 10.54499/LA/P/0083/2020, 10.54499/UIDP/50009/2020, and 10.54499/ UIDB/50009/2020), the Lisbon ELLIS unit, the Center for Responsible AI and FCT PhD grant ref. 2021.09006.BD.
}
}

\author{\IEEEauthorblockN{João Damião Almeida}
\IEEEauthorblockA{\textit{VisLab, Institute for Systems and Robotics} \\
Instituto Superior Técnico, Universidade de Lisboa, Portugal \\
joao.damiao@tecnico.ulisboa.pt}
\\
\IEEEauthorblockN{Cecilia Laschi}
\IEEEauthorblockA{\textit{Soft Robotics Lab} \\
National University of Singapore, Singapore \\
mpeclc@nus.edu.sg}
\and
\IEEEauthorblockN{Egidio Falotico}
\IEEEauthorblockA{\textit{BRAIR Lab, BioRobotics Institute} \\
Scuola Superiore Sant'Anna, Pisa, Italia \\
egidio.falotico@santannapisa.it}
\\
\IEEEauthorblockN{José Santos-Victor}
\IEEEauthorblockA{\textit{VisLab, Institute for Systems and Robotics} \\
Instituto Superior Técnico, Universidade de Lisboa, Portugal \\
jasv@isr.tecnico.ulisboa.pt}
}

\maketitle
\marginnote{\rotatebox[origin=c]{270}{\Huge\color{lightgray}Accepted paper in ICNSC 2025}}

\begin{abstract}
In-hand manipulation tasks, particularly in human-inspired robotic systems, must rely on distributed tactile sensing to achieve precise control across a wide variety of tasks. However, the optimal configuration of this network of sensors is a complex problem, and while the fingertips are a common choice for placing sensors, the contribution of tactile information from other regions of the hand is often overlooked. This work investigates the impact of tactile feedback from various regions of the fingers and palm in performing in-hand object reorientation tasks. We analyze how sensory feedback from different parts of the hand influences the robustness of deep reinforcement learning control policies and investigate the relationship between object characteristics and optimal sensor placement. We identify which tactile sensing configurations contribute to improving the efficiency and accuracy of manipulation. Our results provide valuable insights for the design and use of anthropomorphic end-effectors with enhanced manipulation capabilities.
\end{abstract}


\section{Introduction}

The hand is humans' most important interaction component \cite{Li22}. Its dexterity, crucial for cortical development and cognitive superiority~\cite{Bicchi00}, relies heavily on tactile feedback, especially in challenging in-hand manipulation (IHM) tasks~\cite{Kapassov15}. 

In robotics manipulation literature, dexterity is widely defined as the capability of changing an object's position and orientation from a given reference configuration to an arbitrary different one (within the hand workspace)~\cite{Bicchi00}. IHM is a form of dexterous manipulation without external support, relying on coordinated finger movements, force control, and visual or tactile perception. Since purely vision-based IHM is vulnerable to occlusions and environment changes, tactile sensing becomes a crucial source of information for object manipulation and tool usage~\cite{Johansson79, Yousef11, Dahiya13, Li20, Xiong22}. 

Even though tactile sensory data is an essential element for robotic manipulation, technology and research in artificial tactile sensing is not as developed as for other perception modalities~\cite{Kapassov15, Dahiya13}. Vision-based tactile sensors offer promising capabilities, but still face challenges inherent to optical systems, including vulnerability to lighting conditions, occlusions and surface transparency. While distributed tactile approaches hold promise for robust manipulation skills, they still face many challenges, and sensor placement remains largely limited to the fingertips. We still fail to fully understand the relative importance of the different parts of a hand in tactile perception. 

\begin{figure}[t]
    \centering
    \includegraphics[scale=0.335]{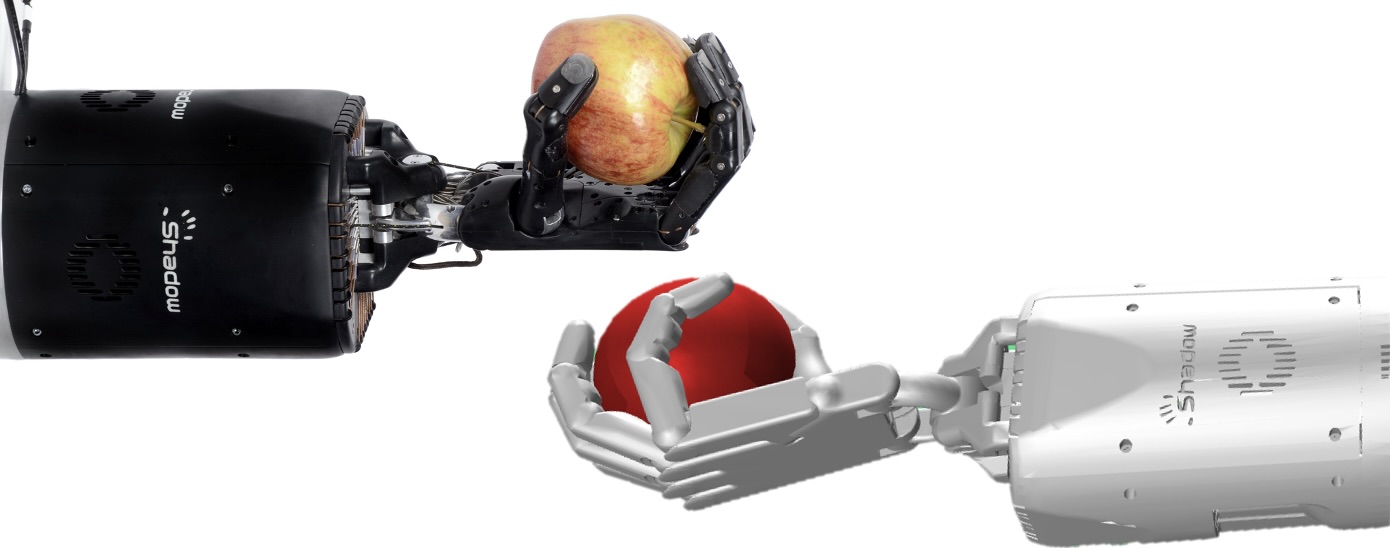}
    \caption {Shadow Hand holding an apple (Photo by ©ESA) and the model of the Shadow Hand holding a sphere, in simulation.}
    \label{intro_shadowhand}
\end{figure}

Reinforcement Learning (RL) enables robots to learn complex motor tasks via feedback-driven exploration ~\cite{Aguinaco23, Lin19}. RL optimizes control policies based on sensory feedback, such as tactile data, that reflects object state. This popular learning approach also helps to evaluate generalization across objects, making it a powerful tool for developing adaptable robotic systems \cite{Itzhak24}. In this study, we train a robotic hand in an Isaac Gym~\cite{Makoviychuk21} RL environment~\cite{Plappert18} (Figure \ref{intro_shadowhand}), leveraging optimized tactile sensor data to improve performance and convergence rate, leading to more efficient object IHM strategies.

We aim to study the relative importance of different hand regions for dexterous IHM and find superior tactile configurations for anthropomorphic hands. To this end, we (i) extend the Isaac Gym RL Shadow-Dexterous-Hand environment with force-torque sensors in 19 areas of the palm and fingers, and (ii) train an agent for object manipulation with different tactile configurations, using a Monte Carlo-inspired approach, to compare sensor placement strategies. We observe that certain tactile regions of the hand contribute more significantly than others, which varies depending on the object size, shape and number of sensors. These results can guide robotics engineers, manufacturers and researchers to design anthropomorphic hands with tactile configurations tailored for IHM.

\section{Related Work}

\subsection{Biological Inspiration} 

\begin{figure}[t]
    \centering
    \includegraphics[scale=0.215]{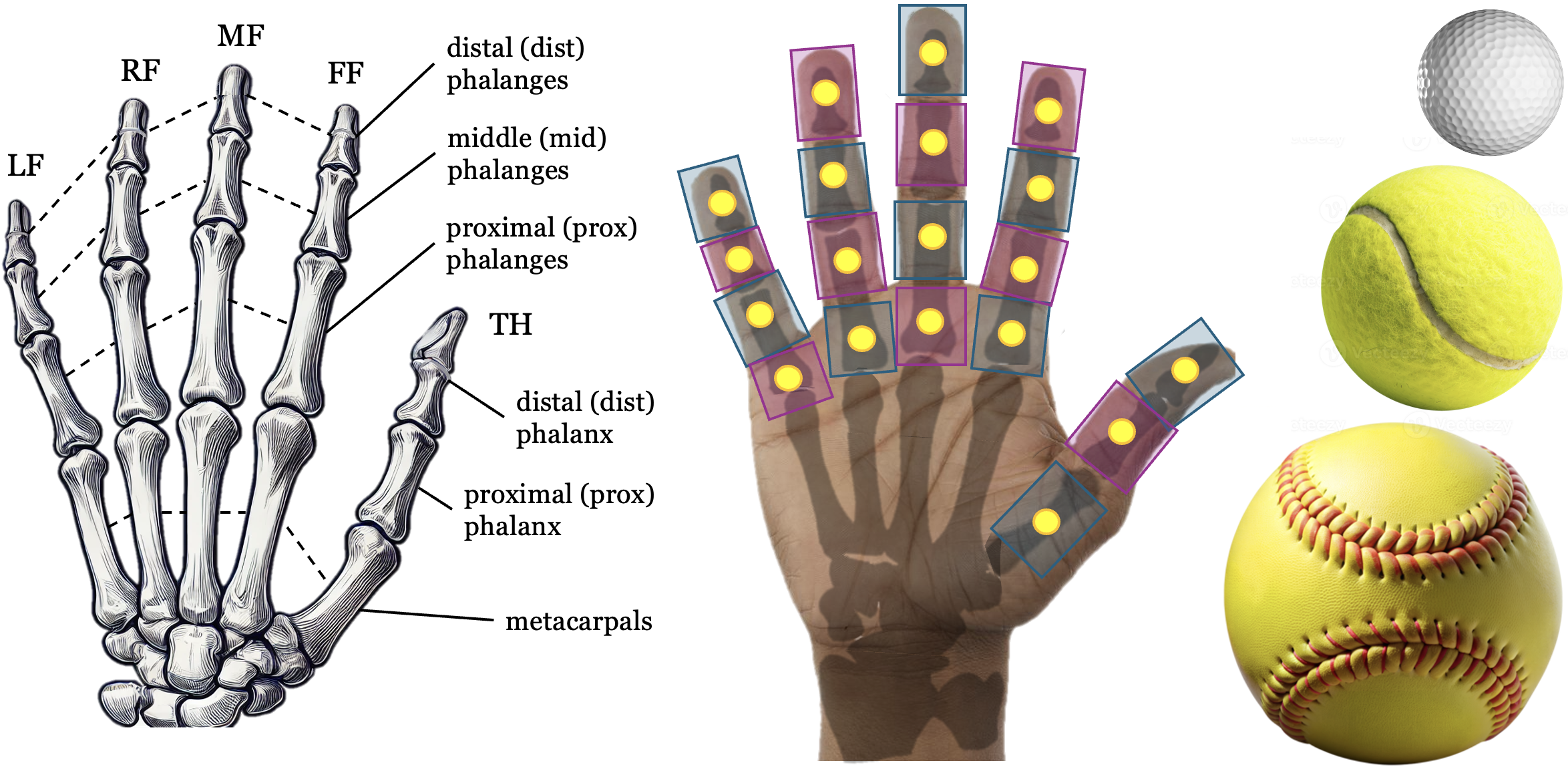}
    \caption {Skeletal anatomy of the human hand (left), approximate contour and center of the 19 sensing regions (middle) and  spherical objects and their relative sizes (right).}
    \label{shadow_anatomy_and_balls}
\end{figure}

The musculoskeletal anatomy of the hand and wrist (Figure \ref{shadow_anatomy_and_balls}), with its 27 DoFs, is one of the most complex systems of the human body. Human hands are also sophisticated \textit{sensory} systems given their anatomical structure and high nerve supply~\cite{Corniani20, Li22}. Tactile sensing, as part of our external haptic system, relies on skin stimulation through mechanoreceptors detecting pressure, vibration, texture or temperature, with the fingertips having the highest concentration of these units~\cite{Johansson79}.

More recently, studies on real individuals~\cite{Sundaram19, Cepria21} wearing sensorized gloves have shown interesting patterns in hand tactile signatures for household activities or object recognition, usually focusing on power grasps and for prosthetics applications. While these anatomical studies are great baselines for replicating our superior dexterity on artificial hands~\cite{Bicchi00}, there is no guarantee that our tactile hand configuration is optimal~\cite{Hu18}, particularly for IHM tasks. Some arguments for this include the intra-species morphological variability of the hand, the fact that the hand serves multiple functions beyond manipulation (e.g. exploration, communication), as well as the fact that routine tasks have changed significantly in the 100.000 years that the human hand has had its current form.

\subsection{Learning In-Hand Manipulation}

Roboticists have made significant progress in mimicking human hand dexterity~\cite{Li22}. Notable anthropomorphic designs include the Shadow Dexterous Hand~\cite{Kochan05} and the DEXMART Hand~\cite{Palli14}. Variations include actuation level (iCub Hand~\cite{Schmitz11}), number of fingers (TriFinger~\cite{Wuthrich20}), or soft components (RBO Hand 3~\cite{Puhlmann22}). The Shadow Hand is commonly used in research and advanced applications~\cite{Li22, Melnik21, Akkaya19, OpenAI18, Plappert18, Guo25}.

\textit{Learning} dexterous manipulation in robots has long been a challenge ~\cite{Lin19, Rajeswaran18, Bicchi00}, with IHM gaining greater attention in recent years~\cite{Pitz24, Bhatt21, Liang20}. While early methods relied on analytical model-based approaches, they posed certain assumptions about the objects and controllers, limiting scalability. To overcome this, (deep) RL allows robots to learn through trial and error by directly interacting with the environment and receiving outcome-based feedback, thus offering greater scalability and flexibility ~\cite{Aguinaco23, Rajeswaran18, Lin19}. Learning from observation with imitation learning improves sample efficiency and encourages more natural manipulation behaviors, although it requires extensive and accurate human demonstrations \cite{Itzhak24}. 

Some works using RL and LSTM architectures have shown great dexterity for object reorientation with a Shadow~\cite{Akkaya19, OpenAI18, Plappert18}. By simulating the task in a physics environment with domain randomization they have shown increased transferability to the real robot. Despite these advancements, most IHM methods remain highly dependent on visual inputs~\cite{Bhatt21, Akkaya19, Rajeswaran18} - making them vulnerable to occlusions and lighting conditions, - and still far from human-level adaptability. 

\subsection{Towards Tactile-based Manipulation} 

Tactile sensing is a long-standing research topic. While hand motor capabilities have at least equaled human levels, their tactile sensor density is modest. Distributed tactile sensing for continuous contact is essential for IHM~\cite{Yousef11} and enables robots to manipulate unknown objects more precisely~\cite{Dong21, Kolamuri21, Dimou22} and complete tasks in touch-only setups~\cite{Yin23, Murali18, Li20, Pitz24}.

Tactile systems typically use piezoresistive or capacitive sensors to collect data on force, pressure, torque, and contact location. While extrinsic sensors provide more precise multi-modal interaction knowledge~\cite{Kapassov15}, they are costly - computationally, financially, architecturally,- and often only fingertips are sensorized~\cite{Li20, OpenAI18, Pitz24, Dimou22}, as in SynTouch BioTac, Xela and Digit sensors. Recently, vision-based tactile sensors have been used to estimate 3D force vectors \cite{Shahidzadeh25} beyond the fingertips \cite{DigitPlexus}. External skin-based tactile systems are an alternative~\cite{Buscher15, Hirai18, Egli24}, though their use in IHM is limited. In both cases, for both developers and researchers, choices and compromises must be made when sensorizing the hand, and literature on this topic is crucial to support decision-making.

While some works confirm the positive role of tactile sensing in robotic manipulation~\cite{Li22, Melnik21, Kolamuri21, Dong21, Liang20, Xiong22}, they generally do not explore \textit{how} this is achieved. Some works have addressed the impact of resolution and binary contact sensors~\cite{Yin23, Liang20, Lin19, Li20, Guo25}. Melnik et al. \cite{Melnik21} map the hand according to the number of contacts. The relative importance of each sensing area and the use of sensors beyond the fingertips are often overlooked~\cite{Li20}, creating a significant gap in current research that our work aims to bridge.

\section{Methodology}

When performing IHM, the hand seeks to adjust the object's position without releasing it (or using controlled tossing and flipping), while the object may slide or roll in the hand. Due to these complex motion patterns, explicit tactile feedback is necessary to infer the current state of the object and achieve safe, effective manipulation. To test dexterous IHM, the task of re-orienting and re-positioning a round object was selected. 

\subsection{Task Description and Metrics}

Object re-positioning and -orientation is a typical yet challenging IHM task. Since irregularities in the shape of the object can influence the use of specific parts of the hand \cite{Cepria21, Sundaram19}, we use a sphere by default, the most symmetrical geometry, for a balanced engagement of the hand's parts during manipulation. We vary the size, mass and mass distribution of the spherical object. As metrics, we consider both accuracy and speed. The task success is defined as achieving the correct orientation and positioning of the object (within a 0.1-radian margin), before the episode ends and without dropping it. 

We formulate the reward function to optimize performance and efficiency. The reward $R$ seeks to minimize distance $d$ (L2 norm) and disalignment $r$ to target, as well as the actions $\mathbf{a}$ required for it:

\begin{equation}
\label{equation_reward}
R = - \alpha d - \beta r - \gamma \|\mathbf{a}\|^2 + S - F
\end{equation}

where $\alpha$, $\beta$, $\gamma$ are scaling factors for distance, rotation, and action penalty; $S$ and $F$ are a bonus and a penalty for task completion or failure, respectively. The metrics used for comparison include the average number of consecutive tasks completed before failure (as in ~\cite{OpenAI18}) and comparative task success rates. Moreover, we measure convergence speed as the number of epochs required for a configuration to achieve baseline performance, as a percentage of its training duration.

\subsection{Hand Sensorial Configuration }

The  artificial anthropomorphic hand chosen to perform this test was the Shadow Dexterous Hand~\cite{Kochan05}, given its similarity to the human hand and its relevance in robotics research. The Shadow Hand has 20 degrees of actuation. The distal phalanges are passively coupled with the middle ones, which mimics human biomechanics.

For sensorizing the hand palm surface, we divided it into 19 regions, corresponding to the fingers' and thumb's phalanges and five more regions in the upper palm. This division was in accordance with our skeletal and cutaneous anatomy, as well as with commercial tactile systems \cite{PPS, SPI} that also measure forces in the phalanges and upper palm, such as Tekscan \cite{Tekscan} which uses 18 sensing regions. Figure \ref{shadow_anatomy_and_balls} shows the skeleton structure of our hand and the disposition of sensing elements. The forces and torques applied in the
whole corresponding region were considered, not just at single points \cite{Makoviychuk21, Vulin21}. 

For each experiment, we would select some of the 19 regions (e.g. 5, 8, or none) in a random or systematic way to provide tactile information. This approach, involving multiple runs with varied sensor placements, allowed us to extensively evaluate the impact of different tactile sensor distributions.

\subsection{Simulation Setup and System Architecture}

The simulation is set up in a RL virtual environment on NVIDIA Isaac Gym~\cite{Makoviychuk21, Plappert18}, a popular physics engine combining GPU-based acceleration, scalable parallelization, force-sensor support, and domain randomization~\cite{Aguinaco23}. We used 4096 parallel environments per experiment, running for a maximum of 3000 epochs ($\sim$1 hour). The Shadow Hand model used is from the OpenAI Gym robotics environments~\cite{Plappert18}, calibrated in~\cite{OpenAI18}. We enhanced it with 19 sensing sites on specific body sub-parts to measure average 3D force and torque applied in each of these segments' volumes (Figure \ref{shadow_anatomy_and_balls}). Forces are measured at the regions’ center, and torques at the sensor’s local reference frame, similarly to \cite{Melnik21, Yin23, Vulin21}.

The system architecture follows~\cite{OpenAI18}, designed for in-hand object re-orientation. The policy is represented as an LSTM with an additional hidden layer (ReLU activations) after the inputs. It is trained with Proximal Policy Optimization, with actions corresponding to desired joint angles relative to the current ones. The reward given at timestep $t$ is in Equation \ref{equation_reward}. For the baseline environments without touch sensing, the 52-dimensional observation space included the Cartesian position of the fingertips and the object pose and relative-to-target orientation. The space was expanded to accommodate varying number of touch sensors, with zero-padding being applied for the non-tactile baseline. We included Gaussian noise in the object and fingertip position to emulate sensor imprecision.

\section{Experiments and Results}

\subsection{Experimental Setup}

For the studied task, the hand starts open and facing up, in a randomized initial resting position. The object is dropped slightly above it. With randomized initial position and orientation, the object has slight variations in mass and non-uniform mass distributions to promote tactile perception. The target position and orientation is also randomly sampled for each task. This simulated environment allowed us to randomize several other parameters (e.g. noise, object properties) and easily change sensor properties and placement. To see the relative importance of a tactile region, we ran dozens of configuration variations, in a Monte-Carlo-inspired approach.

The object used is a 7cm-diameter sphere, similar in size to a tennis ball (see Figure \ref{shadow_anatomy_and_balls}), an object of moderate dimensions requiring finger and palm coordination. For one experiment, we vary the size of the object, keeping the mass non-uniform. For the last experiment, we vary the shape of the object (ellipsoid and cube), with size similar to the tennis ball. 

\subsection{Baselines}

\begin{figure}[ht]
    \centering
    \includegraphics[scale=0.19]{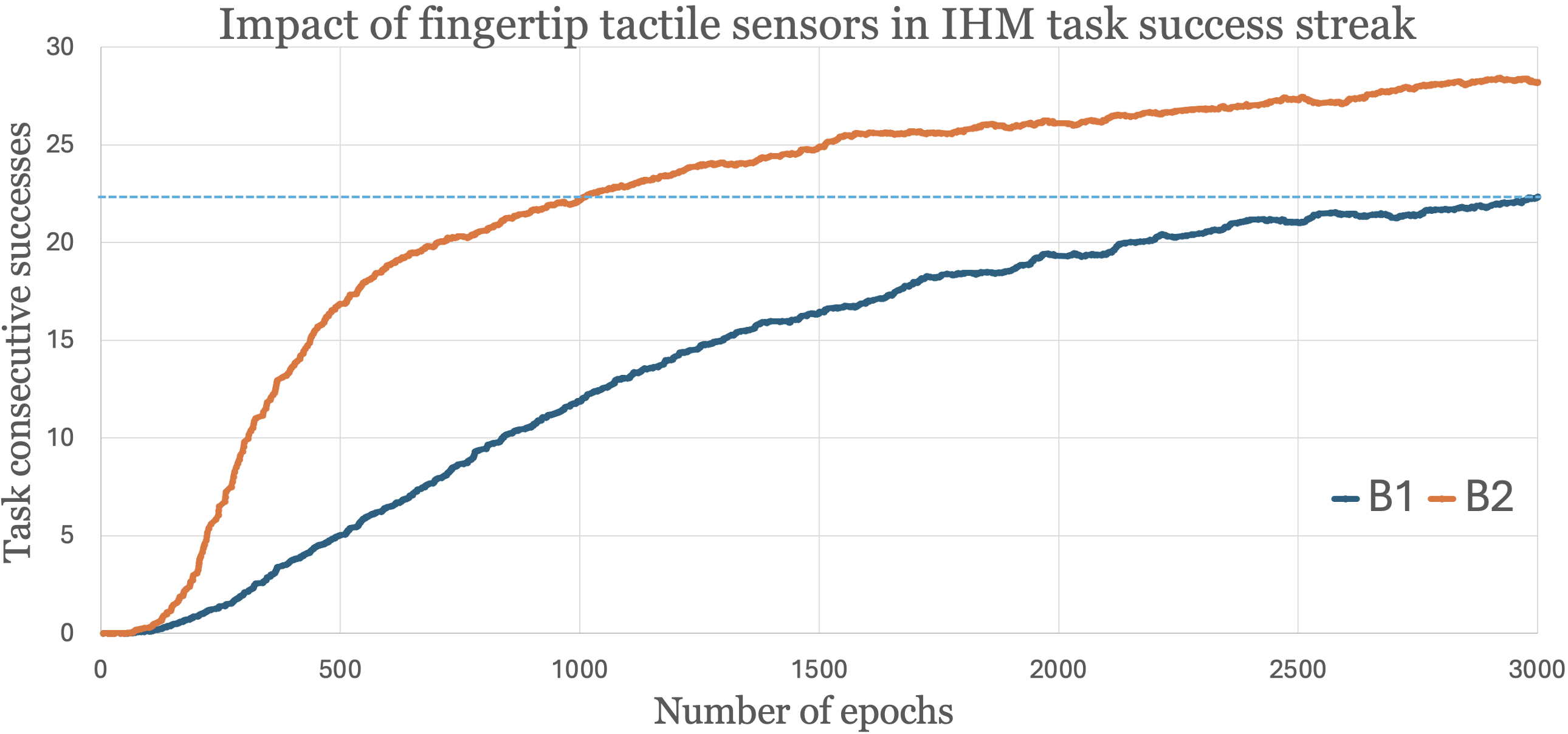}
    \caption {Baselines - consecutive successes: without tactile sensors (B1); with fingertip sensors (B2).}
    \label{baselines}
\end{figure}

The first baseline (B1) performs the task of object re-orientation relying on vision alone. Introduced in~\cite{Plappert18, OpenAI18}, this baseline would already achieve object re-orientation without the use of touch sensors. The second baseline (B2) includes 6D force-torque data from each of the 5 fingertips to perform the same task. For the baselines used, zero-mean Gaussian noise was included to emulate sensor imprecision. The simulation ran for 3000 epochs and the results were smoothed with a moving average of window size 50. Figure \ref{baselines} shows that tactile information improves the success streak up to 26\% and accelerates convergence by a factor of 3.

We also tested this effect with different observational spaces, e.g. featuring joint angles and velocities, as used in~\cite{Plappert18, Melnik21}. Including tactile data always led to better results and faster learning, with the increase being more noticeable for smaller and noisier observational spaces (lower B1 values).

\subsection{Sensor Placement} 

In a first experiment, we moved the (position and touch) sensors from the fingertips (distal phalanges) downwards in the hand: middle phalanges, proximal phalanges, upper-palm. Table \ref{small_table} shows that the fingertip group provided the best results, closely followed by the upper-palm group. The middle and proximal phalanges have similar performance. All increase sample efficiency (2- to 3-fold).

\begin{table}[ht]
\centering
\caption{ Impact of sensor placement latitude in the IHM task }
\label{small_table}
\begin{tabular}{ccccc}
\hline
Exp 1                                                           & \begin{tabular}[c]{@{}c@{}}distal\\ phalanges\end{tabular} & \begin{tabular}[c]{@{}c@{}}middle\\ phalanges\end{tabular} & \begin{tabular}[c]{@{}c@{}}proximal\\ phalanges\end{tabular} & \begin{tabular}[c]{@{}c@{}}upper \\ palm\end{tabular} \\ \hline
\begin{tabular}[c]{@{}c@{}}consecutive \\ successes\end{tabular} & 28.21     & 25.89    & 25.77   & 27.87  \\
\begin{tabular}[c]{@{}c@{}}convergence \\ speed\end{tabular}    & 3.2x    & 2.7x      & 2.2x      & 3.1x      
\end{tabular}
\end{table}

\begin{figure}[ht]
    \centering
    \includegraphics[scale=0.2]{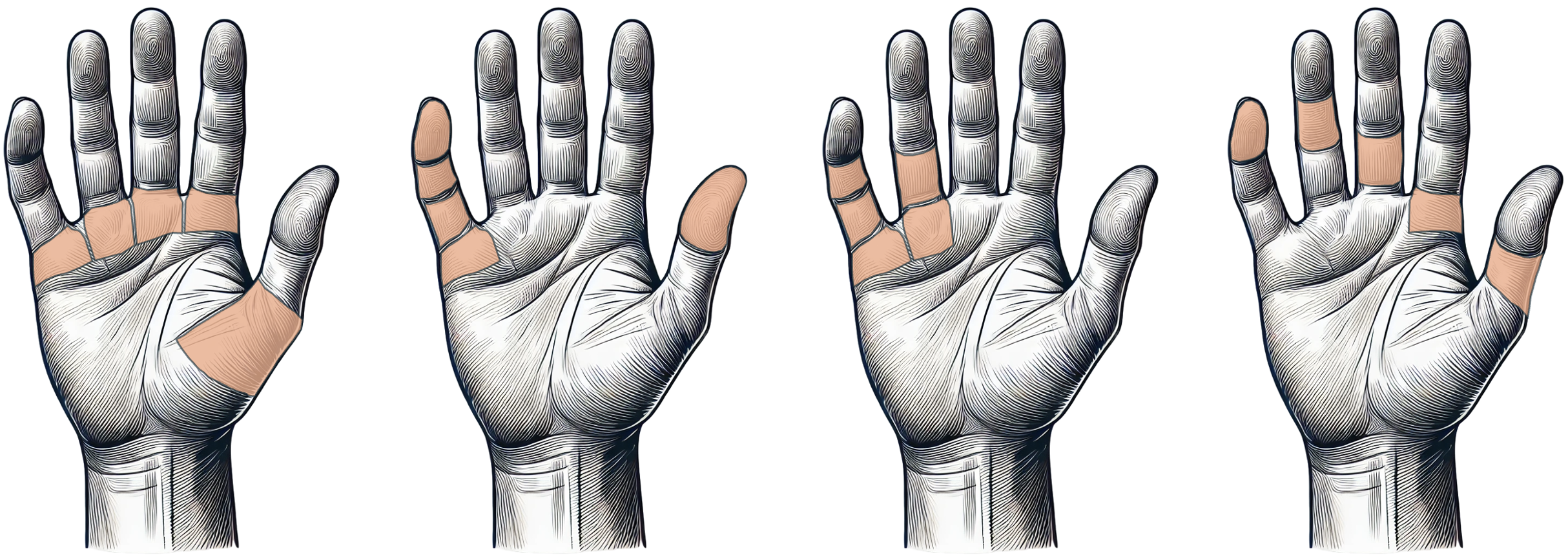}
    \caption {Four configurations from categorical or geometrical sets: same-phalanx (first), same-finger (second), same-”quadrant” (third), diagonally-placed sensors (last).}
    \label{configurations}
\end{figure}

In a second experiment, we relaxed this pre-defined grouping and tested multiple combinations of 5 sensors out of the 19 available locations, so that: 1) all hand regions were tested a similar number of times, 2) each hand region was included in varied configurations with many other sensors, and 3) configurations obeyed some geometrical symmetry or categorical division (see  Figure \ref{configurations}) to allow more interpretable results and for a matter of fairness of comparison. 

The success curves for each configuration were then smoothed (window size of 50 epochs). The average score for each region was calculated and the 19 values were normalized from 0 (worst) to 1 (best). The quantitative relative importance of different phalanges (dist, mid, prox, palm) in different fingers (TH, FF, MF, RF, LF) is presented in Table \ref{big_table} and as a heat map in Figure \ref{heatmaps_exp2and3} (left), for better visualization. The full range corresponds to 18.4\% of the minimum value. The best tested configuration included the  whole thumb and the distal phalanges of the ring and middle fingers.

\begin{table}[b]
\centering
\caption{ Relative importance of the hand's 19 tactile regions }
\label{big_table}
\begin{tabular}{@{}ccccc@{}}
\toprule
Experiments & \begin{tabular}[c]{@{}l@{}}5 sensors \\ (tennis)\end{tabular} & \begin{tabular}[c]{@{}l@{}}5+3 sensors \\ (tennis)\end{tabular} & \begin{tabular}[c]{@{}l@{}}5 sensors \\ (softball)\end{tabular} & \begin{tabular}[c]{@{}l@{}}5 sensors \\   (golf)\end{tabular} \\ \midrule
THdis       & 0.813     & -            & 0.870      & 0.477       \\
THprox      & 0.391     & 0.688        & 0.987      & 0.457      \\
THpalm      & 1.000     & 0.610        & 0.421      & 0.085      \\
FFdis       & 0.612     & -            & 0.923      & 0.219       \\
FFmid       & 0.000     & 1.000        & 0.892      & 0.285       \\
FFprox      & 0.693     & 0.730        & 0.835      & 0.429       \\
FFpalm      & 0.484     & 0.279        & 0.573      & 0.788       \\
MFdist      & 0.610     & -            & 1.000      & 0.430        \\
MFmid       & 0.106     & 0.557        & 0.869      & 0.428        \\
MFprox      & 0.351     & 0.645        & 0.812      & 0.641        \\
MFpalm      & 0.833     & 0.275        & 0.550      & 1.000       \\
RFdist      & 0.882     & -            & 0.596      & 0.000       \\
RFmid       & 0.047     & 0.485        & 0.331      & 0.105       \\
RFprox      & 0.471     & 0.208        & 0.261      & 0.107       \\
RFpalm      & 0.568     & 0.000        & 0.000      & 0.466       \\
LFdist      & 0.643     & -            & 0.564      & 0.141       \\
LFmid       & 0.672     & 0.363        & 0.534      & 0.227       \\
LFprox      & 0.197     & 0.575        & 0.464      & 0.248       \\
LFpalm      & 0.676     & 0.747        & 0.203      & 0.607       \\ \bottomrule 
\end{tabular}
\end{table}

\subsection{Beyond the fingertips} 

Even if it is arguable that the optimal positioning of five sensors should include all five fingertips for IHM tasks, they remain essential for precision tasks or fine manipulation. For many applications, placing sensors on the fingertips is indispensable but adding more sensors is desired. 

\begin{figure}[t]
    \centering
    \includegraphics[scale=0.22]{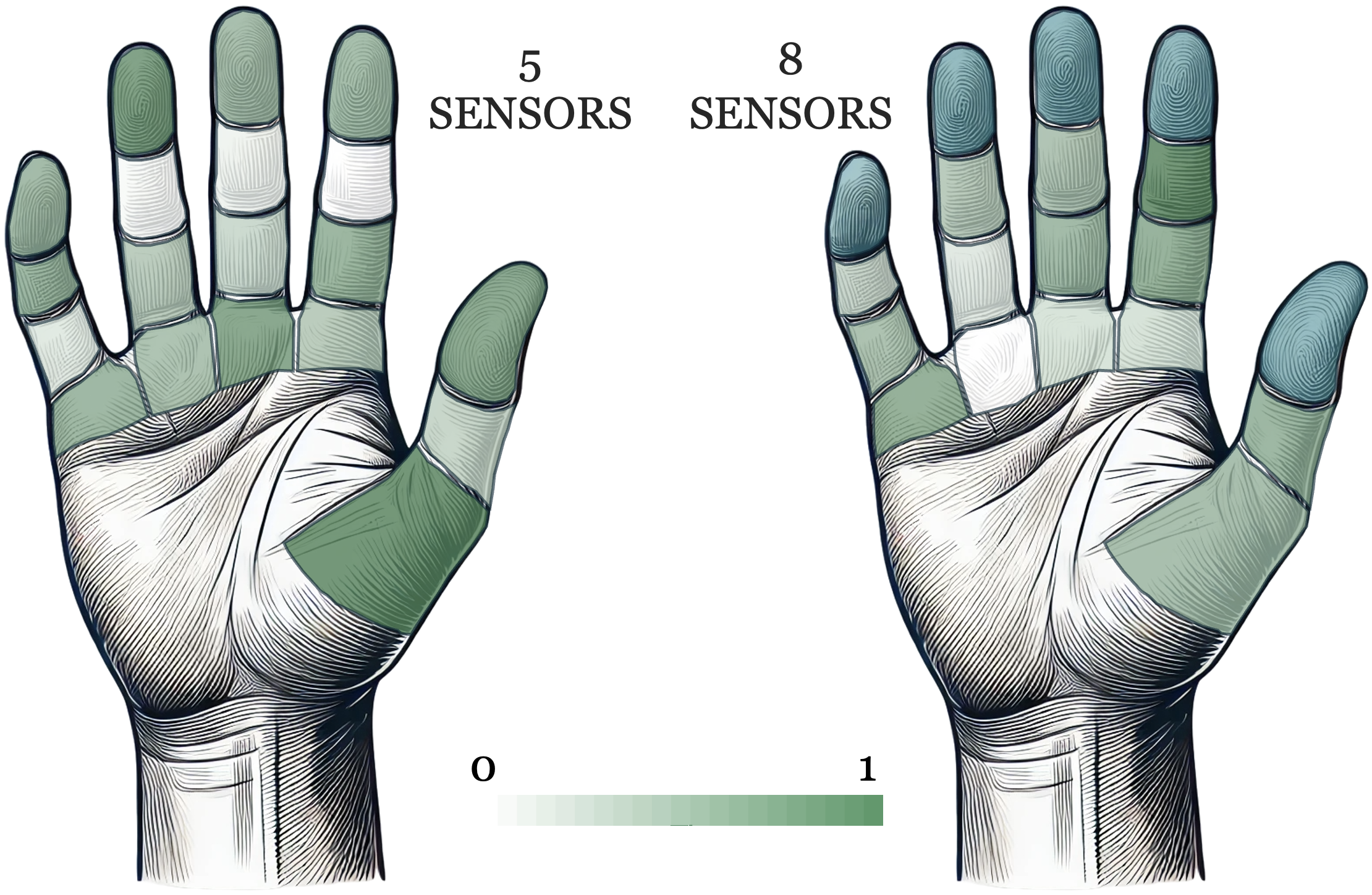}
    \caption {Relative importance (0: Least, 1: Most) of the anthropomorphic hand regions in a IHM task, including the fingertips (left), and besides the fingertips (right).}
    \label{heatmaps_exp2and3}
\end{figure}

In this experiment, we aim to study which parts of the hand, \textit{alongside with} the fingertips, contribute most for this task. We expand the number of sensors to 8 (as found in robotic hands \cite{Aguinaco23, Li22} and tactile gloves \cite{HOGGAN}) and vary the position of 3 of them in the 14 remaining spots. As in the previous experiment, all possible locations feature in multiple tested configurations. The heat map is presented in Figure \ref{heatmaps_exp2and3} (right) and Table \ref{big_table}. The range spanned a 16.9\% increase from the minimum value. 

The results differ noticeably from the 5-sensor configuration. The highest scoring region was the middle phalanx of the FF, followed by its proximal phalanx and the LF base. The highest scoring configuration included thumb's base (palm), and the mid phalanges of FF and RF.

\subsection{Varying object size} 

\begin{figure}[ht]
    \centering
    \includegraphics[scale=0.21]{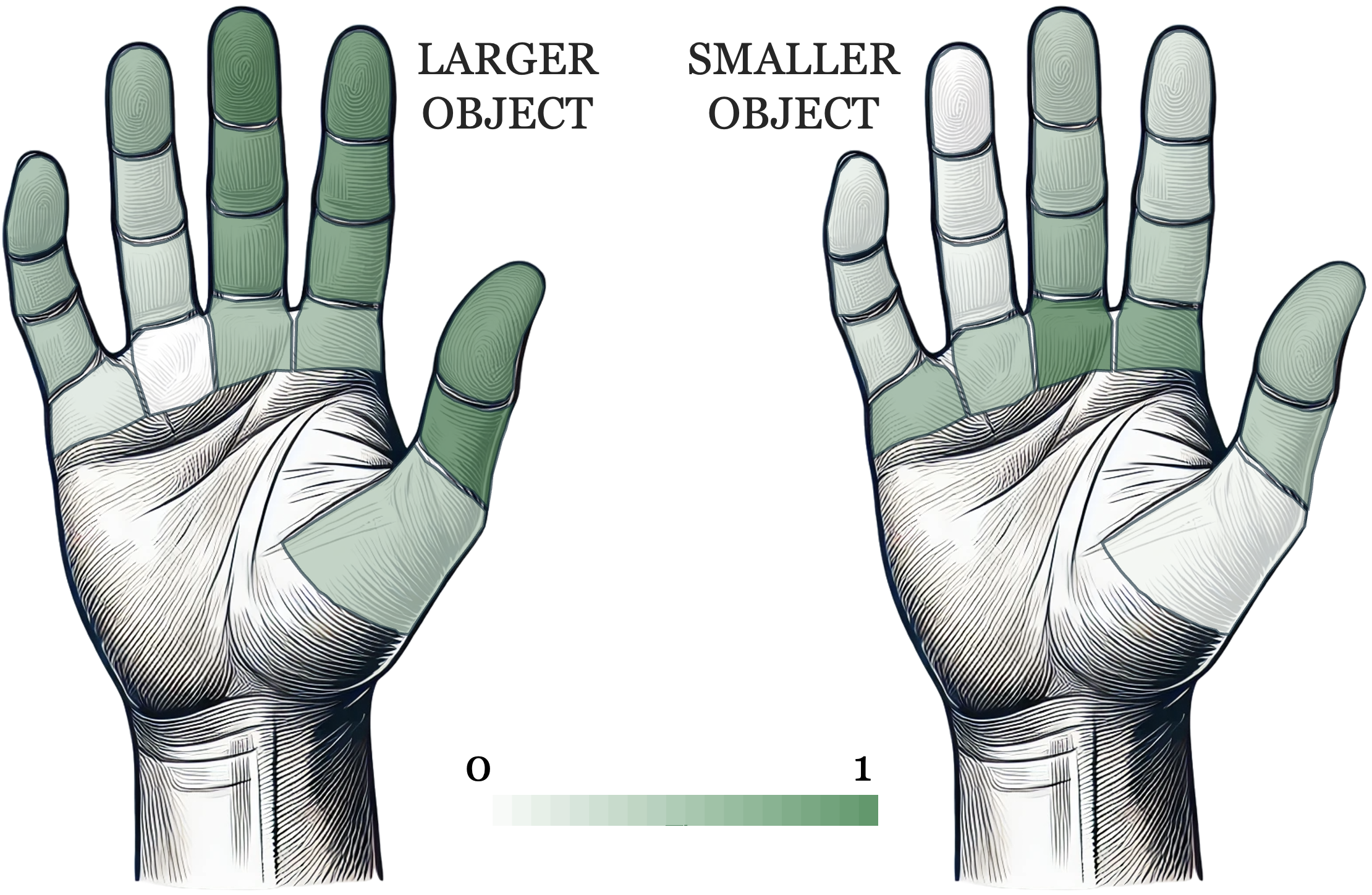}
    \caption {Relative importance (0: Least, 1: Most) of the hand regions in IHM, for larger (left), and smaller objects (right).}
    \label{heatmap_exp4}
\end{figure}

Having used a 7-cm-diameter sphere (tennis ball-sized) for IHM, we now test for 10-cm (softball) and 4-cm (golf ball) spheres. For this experiment (5-sensor configurations), the altered objects had larger or smaller mass (similar density), and the mass distribution was non-uniform as before. Table \ref{big_table} shows the relative contribution of each region (from 0 to 1). These values are also represented as heat maps in Figure \ref{heatmap_exp4}. The failure rate for small spheres was around 4\%, similar to previous experiments, while the same rate for larger spheres was 8\%, probably since they are heavier and harder to manipulate. The full range corresponds to 24.2\% of the lowest value for larger objects, and 41.9\% for the smaller objects, meaning the choice of sensors has more impact in the latter.

For larger spheres, more regions showed high values of average score across configurations, with the TH, FF and MF dominating. The two highest scoring configurations were THdis with whole FF or with whole MF. For smaller spheres, the epicenter of importance was the MF base. The configuration with highest score used the five locations in the palm.

\subsection{Varying object shape}

To study the generality of previous results across other symmetries, one last experiment was performed. For an ellipsoid- and a cubic-shaped object (similar volume to tennis ball), we trained the IHM policy and compared the results using four sensor configurations: 1) no force-torque sensor information (B1 baseline), 2) five sensors on the fingertips (B2 baseline), 3) best tested configuration from Experiment 2 with a tennis ball (THdis, THprox, THpalm, RFdist, MFdist), and 4) the five best performing individual sensors at the same experiment (THdis, THpalm, FFprox, MFpalm, RFdist). 

\begin{figure}[ht]
    \centering
    \includegraphics[scale=0.25]{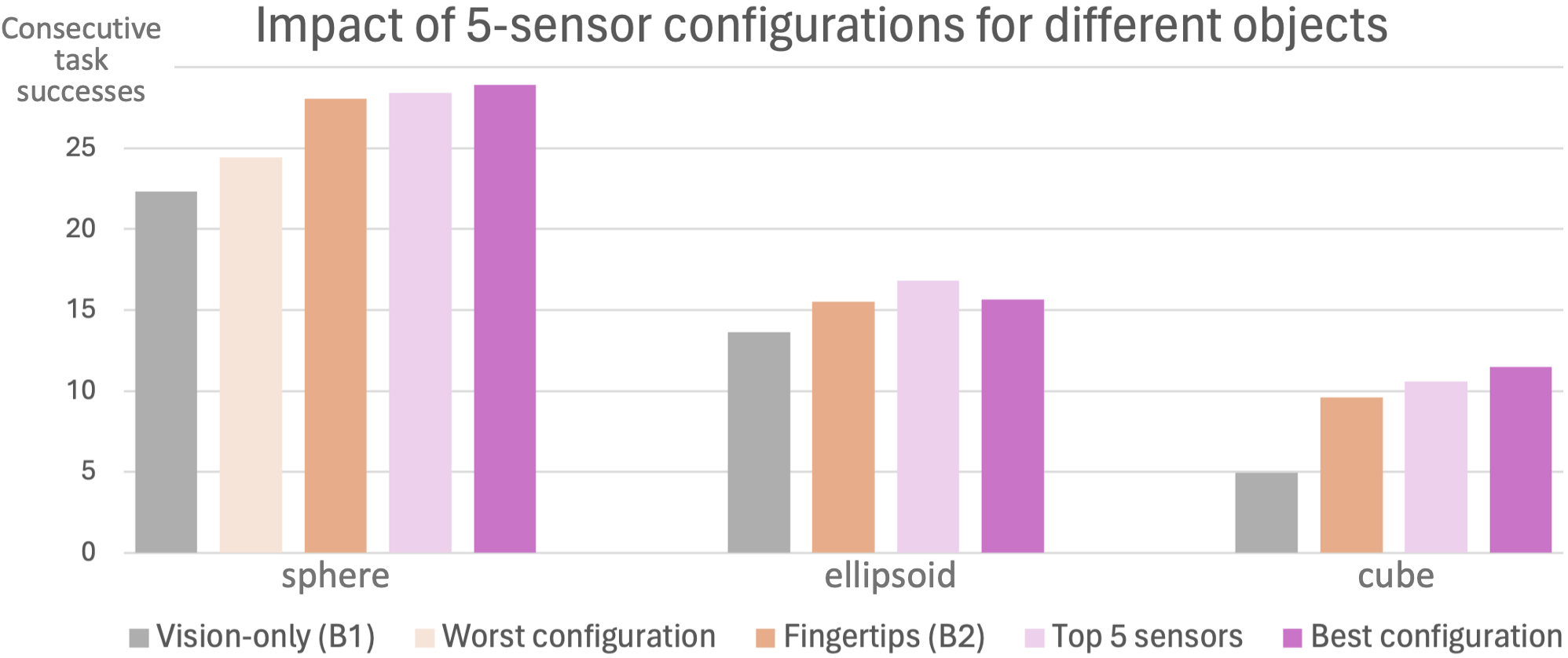}
    \caption {Positive impact of sensors outside of fingertips in spherical, ellipsoidal and cubic object manipulation.}
    \label{different_shape}
\end{figure}

The results are presented in Figure \ref{different_shape} and show that: 1) other shapes are more challenging (decrease in performance), 2) sensory information increases performance for all cases, and 3) the top sensor configurations on spheres still surpass the fingertip configuration (B2) for other shapes.

\section{Discussion and Conclusion}

In this work, we showed that tactile sensing improved learning efficiency and execution consistency, and how the right placement of sensors further improves performance metrics. Overall, the best configurations depended on the task and the object shape and size. The importance of each region is not obvious \textit{a priori}, which shows the relevance of tailoring our design to the type of manipulation.

\subsection{Best tactile configurations} 

We found some results going in line with previous literature on the human hand from a functional perspective~\cite{Cepria21, Sundaram19}. For instance, we notice that the RF, as well as the mid-phalanges, showed on average the worst results and seemed to rarely stand out compared to the other regions. The thumb seemed to always have great importance, whether with its base (mid-size sphere), or tip (others). In addition to the fingertips, the upper palm also showed surprising relative importance. This was less the case in larger objects, which yielded a more distributed contact where fingers took the lead. 

While the human LF is weaker (RF-coupled, smaller muscle support, unrefined control), it may not lack strength in artificial hands and its yaw movement at the hand's extremity~\cite{Zatsiorsky02} gives it surprising relevance. The RF shares some of the drawbacks of the LF, but its position in the hand offers support, enables great pinching amplitude with the thumb, and makes its distal phalanx often relevant. The MF and FF often show similar performance. Depending on the position of the hand and size of the object, they might offer support or perform active rotation, whether at the distal part or the base of the fingers. The thumb plays a pivotal role in the human hand, which translated to its importance here. Its wider range of motion (5 DoF against 3) and its opposable position make it a key element in stabilization, support and dexterity itself.

Fingertips are boundary-phalanges and key in providing precise information about force and object positioning, though their relative importance for IHM is nothing like for precision tasks and grips. Proximal and middle phalanges (both is usually redundant) help in force distribution and tracking object motion when rotating or sliding the object. The upper palm, particularly the thumb's base, provides continuous contact and stability for dynamic manipulation tasks like this.

For complex dynamic movements, well-spread sensor constellations with distributed feedback are crucial for better coverage and continuous contact. The best sensor configuration for these tasks may include distal phalanges but probably not all and definitely not exclusively. Sensor placement is contingent on the total sensor count, as increasing the number does not guarantee that previously optimal locations remain advantageous. In the experiment with 8 sensors, as distributed and stabilizing contact points were already guaranteed with the fingertips, the middle phalanges also gained importance. So, depending on the task and the number of available sensors, a deeper understanding of their placement strategies may help the artificial system learn much faster and with better results.

\subsection{Correlation with size and shape} 

Different sensing regions give important contributions and can integrate effective configurations depending on the object, hand pose and manipulation strategy~\cite{Yousef11, Bicchi00}. The results changed significantly with the variations in object size but showed transferability to other shape primitives.

Smaller objects led to greater margins in the relative importance of hand regions, since less surface area is in contact with the sphere. Larger spherical objects demanded greater contribution from a distributed sensing area, with a more balanced contribution of all hand regions. In general, while the palm and central fingers are important to sense contact, FF and LF are important in balancing large objects and may produce large forces since they are responsible for stabilization and rotation (torque) of objects, according to the mechanical advantage hypothesis ~\cite{Zatsiorsky02}. If fingertips are usually key in manipulating fine objects, here the hand palm stands out for small sphere manipulation, likely due to the upward-facing hand setup, allowing the sphere to fall into it. It was common to see the thumb rolling it against the fingers' bases, in a manipulation technique which is not strange to us.

In human IHM, the shape of the object influences how we grasp and rotate and therefore which sensing regions to use. The results showed that tactile sensing improved vision-only baselines (B1) for all tested geometries. The best configurations for sphere manipulation also surpassed non-optimal fingertip-only constellations for the ellipsoid and the cube. The principles derived from these shapes can be generalized. As pointed out in \cite{Sartori11}, we tend to grasp and manipulate objects by their thinner axis, to maximize stability. This indicates that the size of the minimum axis might play a larger role in tactile use than the shape of the object itself, which our results confirmed.

\subsection{Key Takeaways}
In general, for the task of in-hand object manipulation:
\begin{itemize}
    \item Tactile sensing improved vision-only baselines for all tested geometries.
    \item Well-spread sensor constellations with distributed feedback are key for better coverage and continuous contact.
    \item The best sensor configuration for these tasks includes distal phalanges but not all of them and not exclusively.
    \item The optimal network of sensors depends on their number: adding more sensors doesn’t always mean placing them where they were previously useful. 
    \item Results changed significantly with the variations in object size but showed transferability to other shape primitives.
\end{itemize}

And more specifically, regarding their tactile importance:
\begin{itemize}
    \item Middle finger base, thumb tip and middle finger tip consistently showed positive impact across object sizes.
    \item Ring finger middle phalanx showed the worst results on average across different object sizes and shapes.
    \item If possible to add more sensors besides fingertips for similar tasks, sensitizing the other phalanges of the first finger proved to be high-priority.
\end{itemize}

\section{Scope and Future Work} 
\label{sec:future_work}

This work focuses on a specific anthropomorphic robotic hand reorienting and repositioning symmetrical geometries with non-uniform mass. The methodological choices in this complex, multi-variable optimization problem were carefully justified, and the findings may offer valuable insights. The research problem of optimal tactile sensorization has a broad scope and future work could involve expanding these results for other tasks, tools, manipulators or learning methods.

Despite high-fidelity simulation, our model abstracts real-world sensor noise and the precise, often deformable, contact mechanics of tactile sensors, which present inherent challenges for direct physical deployment. Future work could improve sensor model realism. Nevertheless, while sim-to-real transfer is beyond the scope of this work, our design choices were intentionally oriented toward transferability. We build on the OpenAI framework~\cite{OpenAI18}, which trained a Shadow Hand in simulation and successfully deployed it for cube rotation using fingertip sensors. We used the same simulation tool, Mujoco models, similar observational space, and calibrated parameters from physical robot’s available data~\cite{Plappert18}. The transfer in~\cite{OpenAI18} was achieved by extensive domain randomization (besides large-scale training with distributed RL), which we also kept in our implementation. Since this work compares many sensory configurations and their relative impact, a simulated environment was considered most fitting. Tests on a real Shadow Hand are possible using tactile sensors \cite{DigitPlexus, Tekscan} or gloves \cite{PPS, HOGGAN}. 

This topic offers other opportunities for future research. The proximal and interior parts of the hand palm were not sensorized in our current setup. Our study focuses on the right hand and it would be valuable to extend it for bi-manual tasks. It could also be interesting to test different characteristics of the sensors \cite{Akinola24} - resolution, force range, noise tolerance, - and compare with the tactile diversity of our own human hand.

\end{document}